\pdfoutput=1

\documentclass[11pt]{article}

\usepackage[preprint]{acl}

\usepackage{times}
\usepackage{latexsym}

\usepackage[T1]{fontenc}

\usepackage[utf8]{inputenc}

\usepackage{microtype}

\usepackage{inconsolata}

\usepackage{graphicx}

\usepackage{caption}
\usepackage{booktabs}
\usepackage{subcaption}
\usepackage{multirow}
\usepackage{hyperref}
\usepackage{cleveref}

\title{Leveraging Parameter-Efficient Transfer Learning for Multi-Lingual Text-to-Speech Adaptation}

\author{
 \textbf{Yingting Li \textsuperscript{1}},
 \textbf{Ambuj Mehrish \textsuperscript{2}},
 \textbf{Bryan Chew \textsuperscript{2}},
 \textbf{Bo Cheng \textsuperscript{1}},
 \textbf{Soujanya Poria \textsuperscript{2}},
\\
 \textsuperscript{1}Beijing University of Posts and Telecommunications,\\
 \textsuperscript{2}Singapore University of Technology and Design,
\\
\texttt{\{cindyyting, chengbo\}@bupt.edu.cn}
\{bryan\_chew\}@mymail.sutd.edu.sg\\
\{ambuj\_mehrish, sporia\}@sutd.edu.sg\}
}

\begin{document}
\maketitle
\begin{abstract}
Different languages have distinct phonetic systems and vary in their prosodic features making it challenging to develop a Text-to-Speech (TTS) model that can effectively synthesise speech in multilingual settings. Furthermore, TTS architecture needs to be both efficient enough to capture nuances in multiple languages and efficient enough to be practical for deployment. The standard approach is to build transformer based model such as SpeechT5 and train it on large multilingual dataset. As the size of these models grow the conventional fine-tuning for adapting these model becomes impractical due to heavy computational cost. In this paper, we proposes to integrate parameter-efficient transfer learning (PETL) methods such as adapters and hypernetwork with TTS architecture for multilingual speech synthesis. Notably, in our experiments PETL methods able to achieve comparable or even better performance compared to full fine-tuning with only $\sim$2.5\% tunable parameters\footnote{The code and samples are available at:~\url{ https://anonymous.4open.science/r/multilingualTTS-BA4C}}.


\end{abstract}

\section{Introduction}
Multilingual speech synthesis, generating speech in multiple languages from text input, represents a major advancement in speech processing with wide-reaching implications for global communication~\cite{tan2021survey,mehrish2023review}. Unlike single-language systems, multilingual architectures break linguistic barriers, transforming education, entertainment, healthcare, and customer service by facilitating seamless communication across languages~\cite{marais2020awezamed,seong2021multilingual,le2024voicebox,panda2020survey}.

Current multilingual TTS architectures face challenges \cite{nuthakki2023deep,kaur2023conventional}, including the complexity of modeling diverse linguistic structures, phonetic variations, and prosodic features across languages. Resource constraints, such as the availability of multilingual corpora and linguistic expertise, can impede model development, particularly for low-resource languages or underrepresented dialects \cite{tan2021survey,mehrish2023review}. Addressing these challenges requires concerted efforts in data collection, model development, and evaluation.
\begin{figure*}[ht]
    \centering
    \includegraphics[width=2\columnwidth]{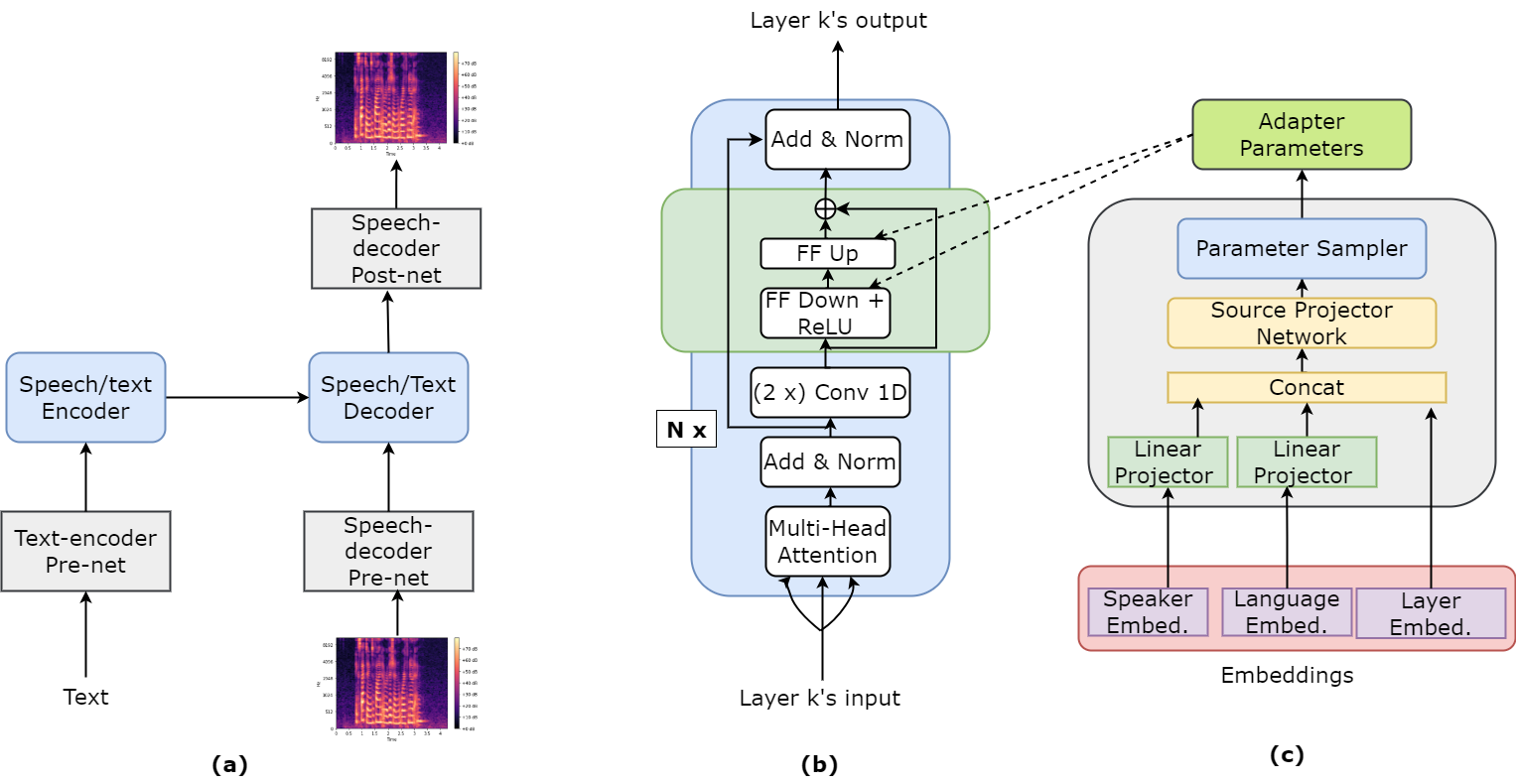}
    \captionsetup{font=small} 
    \caption{(a) SpeechT5 TTS architecture. (b) Encoder/decoder architecture with adapter. (c) HyperGenrator architecture.}
    \label{fig:speechT5}
\end{figure*}

The advancement of architectural designs, coupled with pre-training models such as SpeechT5 \cite{ao2021speecht5}, reflects challenges similar to those encountered in NLP. Achieving optimal performance through fine-tuning these models for diverse downstream tasks or domain adaptations requires substantial task-specific datasets. Moreover, fine-tuning all model parameters necessitates significant memory resources allocated to each task. With limited data available for various underrepresented languages, full fine-tuning can further leads to poor generalization. Researchers have sought solutions to these challenges through the exploration of PETL methods \cite{li2023evaluating, oh2023blackvip, chen2023efficient, 9746126, vanderreydt2023parameter, le2021lightweight}. However, their investigation remains limited for TTS adaptation.

In this paper, we extends PETL approaches to the  multilingual TTS, focusing on adapter \cite{houlsby2019parameter} and Hyper-Network \cite{ustun2022hyper}. We pioneer the hyper-networks for multilingual TTS adaptation and introduces the \textit{Multi-Conditioned HyperGenerator} for multilingual TTS. Our major contributions includes: (1) Regular \& Dynamic Adapters: We embed language-specific parameters into SpeechT5 using regular adapters and explore a hyper-network to generate these parameters, referred to as HyperGenerator. (2) Parameter Efficiency: We achieve comparable or superior performance to full fine-tuning using only about $2.44\%$ of the parameters. (3) Improved Zero-shot Performance: HyperGenerator outperforms full finetuning and regular adapters on an unseen language with the same parameter count.


\section{Related work}
Research on PETL methods for large multilingual pretrained models like XLSR \cite{vanderreydt2023parameter} has gained significant attention, notably in Automatic Speech Recognition (ASR) \cite{fu2023effectiveness, zhang2023google, yu2023low, yang2023english, shi2024exploration}. Li et al. \cite{li2023evaluating} propose a benchmark utilizing XLSR-53 \cite{conneau2020unsupervised}, employing PETL such as regular adapters \cite{pfeiffer2020adapterhub}, prefix tuning \cite{li2021prefix}, and LoRA \cite{hu2021lora}. Le et al. \cite{le2021lightweight} and Zhao et al. \cite{zhao2022m} explore multilingual neural machine translation, focusing on lightweight adapter tuning. Morioka et al. \cite{morioka2022residual} advocate for integrating regular adapters with TTS models for few-shot speaker adaptation, while Mehrish et al. \cite{mehrish2023adaptermix} introduce a mixture of experts for low-resource speaker adaptation.






\section{Methodology}

\subsection{Base Model Architecture: SpeechT5}
SpeechT5 \cite{ao2021speecht5} merges NLP and speech synthesis techniques, extending the transformer-based T5 architecture~\cite{raffel2020exploring}. It integrates self-attention mechanisms and CNNs to capture both temporal dependencies and spectral features in speech. By pre-training on large-scale speech corpora and fine-tuning on specific datasets, SpeechT5 excels in tasks such as speech recognition, TTS, and speech translation. 

\subsection{Adapter}
In this work, we integrate language-specific parameters using adapter modules, commonly employed in the NLP for multilingual or multi-task scenarios~\cite{ansell-etal-2021-mad-g}. Following the formulation of \cite{houlsby2019parameter}, we insert one adapter block after each convolutional block of every transformer module in the SpeechT5 model as shown in Figure \ref{fig:speechT5}. Each adapter module, with fewer parameters compared to the main network (SpeechT5), down-projects the input to a lower-dimensional space, applies a non-linearity, and then up-projects back to the original dimensions. A residual connection is added to produce the final output. During language adaptation, only the adapter parameters are updated while keeping the main network frozen.
\begin{table*}[h!]
\centering
\resizebox{\textwidth}{!}{
\begin{tabular}{lccccccccccccl}
\toprule 
 \multirow{2}{4em}{Model} & \multicolumn{2}{|c|}{de} & \multicolumn{2}{|c|}{fr} & \multicolumn{2}{|c|}{fi} &  \multicolumn{2}{|c|}{hu} & \multicolumn{2}{|c|}{nl } & \multicolumn{2}{|c|}{avg} &  \multirow{2}{4em}{\textbf{Params}}  \\
 &  MCD & CER  & MCD & CER & MCD & CER & MCD & CER &  MCD & CER  &  MCD & CER  &  \\ 
\midrule
Finetune Multilingual w/o Pretrain & $4.88$ & $7.19$ & $4.91$ & $14.40$ & $4.76$ & $15.16$ & $4.93$ & $16.29$ & $5.46$ & $10.56$ & $4.99$ & $12.72$ & 144M(100\%) \\ 
\midrule
Finetune Multilingual w/ Pretrain & $4.87$ & $6.66$ & $4.95$ & $11.82$ & $4.82$ & $12.64$ & $5.00$ & $15.52$ & $5.51$ & $10.28$ & $5.03$ & $11.38$ & 144M(100\%) \\
Adapter Multilingual w/ Pretrain & $4.75$ & $\textbf{6.30}$ & $4.93$ & $11.81$ & $4.75$ & $9.58$ & $4.92$ & $16.11$ & $5.40$ & $10.37$ & $4.95$ & $10.83$  & 3.56M(2.47\%) \\
HyperAdapter Multilingual w/ Pretrain  & $4.78$ & $6.52$ & $5.04$ & $15.50$ & $4.67$ & $\textbf{7.05}$ & $\textbf{4.87}$ & $\textbf{13.13}$ & $\textbf{5.36}$ & $10.96$ & $4.94$ & $\textbf{10.63}$ & 3.52M(2.44\%) \\
\midrule
Finetune Monolingual w/ Pretrain  & $4.86$ & $6.44$ & $4.85$ & $\textbf{10.80}$ & $4.67$ & $15.09$ & $4.90$ & $15.54$ & $5.53$ & $\textbf{8.47}$ & $4.96$ & $11.27$  & 144M(100\%) \\
Adapter Monolingual w/ Pretrain  & $4.77$ & $6.82$ & $4.82$ & $14.32$ & $\textbf{4.63}$ & $9.41$ & $4.94$ & $14.53$ & $5.36$ & $14.22$ & $\textbf{4.90}$ & $11.86$  & 3.56M(2.47\%) \\
HyperAdapter Monolingual w/ Pretrain  & $\textbf{4.71}$ & $7.94$ & $\textbf{4.82}$ & $14.66$ & $4.77$ & $13.47$ & $4.93$ & $14.00$ & $5.40$ & $13.21$ & $4.93$ & $12.66$  & 3.52M(2.44\%) \\        
\bottomrule
\end{tabular}
}
\caption{Evaluation results for\textit{ seen} languages along with the percentage of parameters updated during training.
\label{tab:hypernetwork_ablations_dimension_decoder}
}
\end{table*}

\subsection{HyperGenerator}
HyperGenerator consists of a hyper network  \cite{ustun2022hyper} that generates the weights of all adapter modules. As depicted in \Cref{fig:speechT5}, a single hyper-network is employed to create adapters for multiple languages and layers, with conditioning on $(s,l,p)$, where $s$ denotes speaker embeddings, $l$ represents the target language, and $p$ indicates the encoder or decoder layer ID. This method, unlike traditional adapters, promotes cross-language and cross-layer information sharing, enabling the hyper-network to efficiently distribute its capacity among them. By adapting parameters based on speaker characteristics and language specifics, the hyper-network augments the effectiveness of adapters. Furthermore, to ensure network efficiency, we utilize a shared hyper-network to generate adapter parameters across all layers within the TTS backbone, further conditioning it with the layer ID $p$.

\section{Experimental Setup}
\subsection{Baseline and Dataset}
\label{sec:base}
We developed a baselines with the following $3$ configurations for comparing the performance of the adapter and HyperGenerator with full fine-tuning.

\noindent \textbf{Monolingual}: We finetune the SpeechT5 individually for each language, uniquely optimizing its parameters to enhance speech synthesis performance.

\noindent \textbf{Multilingual}: We finetune the SpeechT5 with diverse speech data from multiple languages. This improves the model's ability to understand and generate speech across various linguistic contexts, capturing cross-lingual patterns, phonetic variations, and language-specific features.
 \begin{figure}
\centering
    \includegraphics[width=\columnwidth]{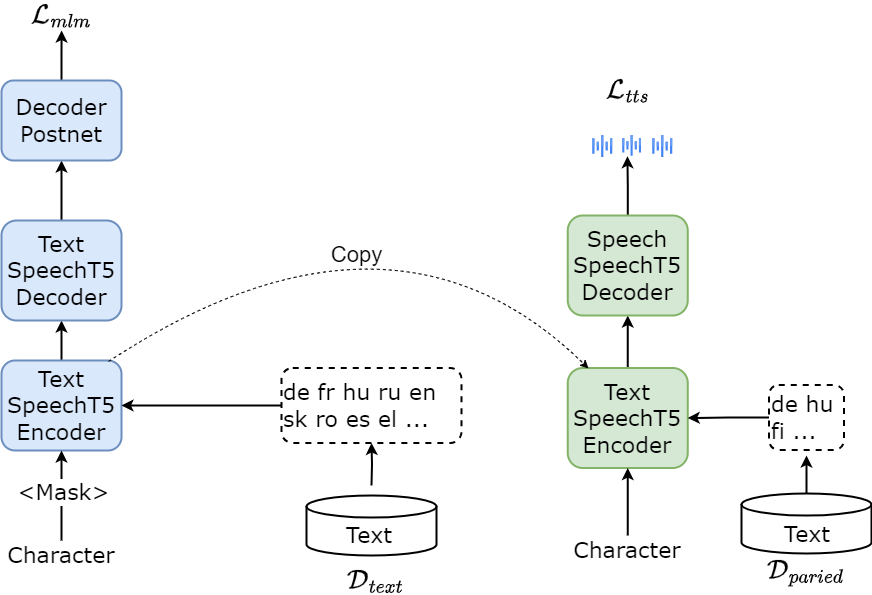}
    \caption{Multilingual Masked Text Pretraining. Where $\mathcal{L}_{mlm}$ and $\mathcal{L}_{tts}$ is mask language modeling and reconstruction loss respectively. }
    \label{fig:masked_parameter_transfer}
\end{figure}

\noindent \textbf{Multilingual Masked Text Pretraining}: Multilingual models like multilingual BERT~\cite{DBLP:journals/corr/abs-1810-04805} have demonstrated strong cross-lingual transfer capabilities in NLP tasks. Leveraging multilingual pre-training improves generalization to other languages without specific target data. In this settings, we extend MLM pre-training to SpeechT5 to enhance pronunciation and prosody transfer. The left side of Figure \ref{fig:masked_parameter_transfer} illustrates the unsupervised pre-training of SpeechT5's text encoder and decoder using text-only data $\mathcal{D}{_{text}}$ with MLM. The pre-trained text encoder is then integrated into the TTS pipeline, as shown on the right side of Figure \ref{fig:masked_parameter_transfer}, and trained on paired speech-text data $\mathcal{D}{_{paired}}$.

\begin{figure*}[ht]
    \centering
    \subfloat[Multilingual performance]{\includegraphics[width=\columnwidth, height=0.7\columnwidth]{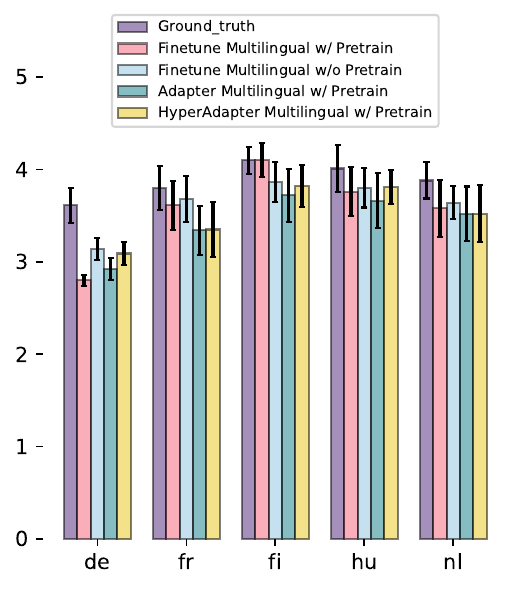}\label{fig:multilingual}}
    \hfill
    \subfloat[Monolingual performance]{\includegraphics[width=\columnwidth, height=0.7\columnwidth]{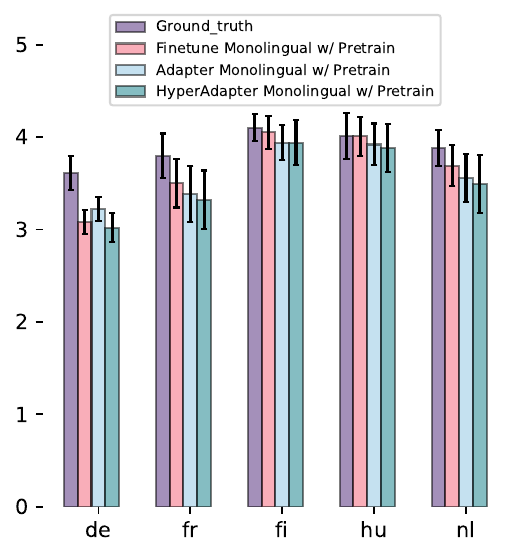}\label{fig:monolingual}}
    \caption{Subjective evaluation on naturalness: MOS score}
    \label{fig:comparison}
\end{figure*}
For fine-tuning using monolingual and multilingual configuration, as discussed in Section \ref{sec:base}, we leverage German (de), French (fr), Finnish (fi), Hungarian (hu), and Dutch (nl)—as the five \textit{seen} European languages from the CSS10 dataset \cite{park2019css10}. To evaluate zero-shot performance, we use Spanish (es) as an \textit{unseen} language. For Multilingual Masked Text Pretraining, we utilize transcripts from VoxPopuli \cite{wang2021voxpopuli}, M-AILABS \cite{bakhturina2021hi}, and CSS10 \cite{park2019css10} to pre-train the SpeechT5 text encoder-decoder for a character-based masked language modeling task. 

\subsection{Training and Evaluation}
 We follow the data partition outlined in \cite{Saeki2023LearningTS}. We use pretrained chekcpoint\footnote{\url{https://huggingface.co/microsoft/speecht5_tts}} of SpeechT5 for all experiments. Speaker embeddings\footnote{Pretrained speaker verification model \cite{wan2018generalized}.} are set at a dimension of $256$, while language embeddings are initialized using pretrained weights from lang2vec \cite{littell2017uriel}. The layer embedding dimension is set at $64$. The bottleneck dimension for adapters is $128$, whereas for HyperGenerator, is $32$ for ensuring the same number of parameters across both architectures. We employed MCD \cite{kominek2008synthesizer} and assess intelligibility using Character Error Rates (CERs) computed with the multilingual ASR \cite{radford2023robust}\footnote{https://github.com/openai/whisper} as objective metrics. Furthermore, to evaluate naturalness, we conducted listening tests to calculate the MOS of synthesized speech. We recruited five native speaker via Amazon Mechanical Turk (AMT) for each of the languages. 
\begin{table}[h!]
\centering
\resizebox{\columnwidth}{!}{
\begin{tabular}{lccc}
\toprule
 \multirow{2}{4em}{Model} & \multicolumn{2}{|c|}{es} \\
 &  MCD & CER   \\ 
\midrule
Finetune Multilingual w/o Pretrain & 5.75 & 39.32 \\
\midrule
Finetune Multilingual w/ Pretrain & 5.87 & 34.80 \\
Adapter Multilingual w/ Pretrain & 5.39 & 45.94 \\
HyperAdapter Multilingual w/ Pretrain  & \textbf{5.28} & \textbf{18.79} \\
\bottomrule
\end{tabular}}
\caption{Evaluation results for \textit{unseen} language (es).
\label{tab:zero_shot_perform}
}
\end{table}

\section{Results}
\subsection{Objective and Subjective Evaluation}
Table \ref{tab:hypernetwork_ablations_dimension_decoder} shows that Finetune \textit{Multilingual with Pretraining} outperforms Finetune \textit{Multilingual without Pretraining} due to multilingual masked text pretraining. Both Adapter and HyperGenerator achieve similar or better performance than full fine-tuning with significantly fewer parameters. Full fine-tuning updates 144M parameters, while Adapter and HyperGenerator use only 3.56M and 3.52M parameters, respectively, with HyperGenerator showing superior performance in multilingual settings with text pretraining.

\begin{figure}[ht]
    \centering
    \includegraphics[width=\columnwidth]{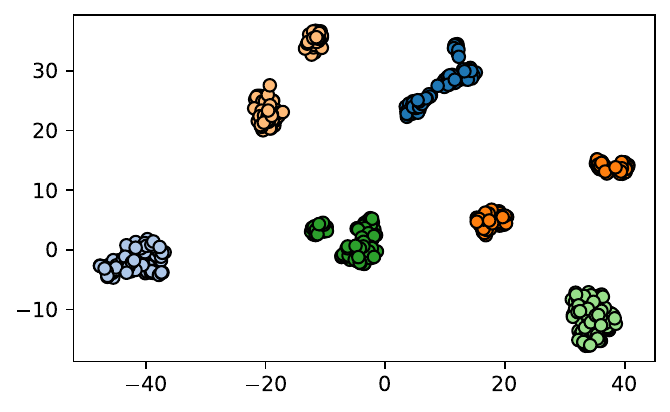}
    \caption{t-SNE plot of HyperGenerator parameters for 6 languages from the CSS10 test set, with same colors denoting speech samples from the same language.}
    \label{fig:hyperx_language_speaker}
\end{figure}

While the performance gain for HyperGenerator with \textit{seen} languages is modest, it shows promise for zero-shot multilingual speech synthesis. Table \ref{tab:zero_shot_perform} shows that HyperGenerator achieved a CER of 18.79\% for Spanish, significantly lower than the over 30\% CER for Multilingual Fine-tuning and adapter-based approaches. This highlights HyperGenerator's dynamic adaptability and potential for efficient and accurate zero-shot synthesis. Figure \ref{fig:hyperx_language_speaker} further demonstrates this, as speech from the same language clusters together, indicating HyperGenerator's ability to adjust parameters based on language, unlike static adapters.
MOS results (\Cref{fig:multilingual} and \Cref{fig:monolingual}) indicate that Adapters and HyperGenerator perform as well as or better than full fine-tuning in multilingual contexts. HyperGenerator consistently achieved the highest scores, with values of 3.09 for \textit{de} and 3.81 for \textit{hu}, demonstrating superior naturalness. Similar trends in monolingual scenarios highlight HyperGenerator's effectiveness in generating high-quality speech across different languages.



\section{Conclusion}
In this paper, we advances multilingual speech synthesis using PETL methods like adapter fine-tuning, achieving SOTA performance with fewer parameters. Introducing regular and dynamic adapters with a hyper-network enhances efficiency and zero-shot performance. Future work could optimize adapters for specific languages, improve cross-lingual transfer learning, and reduce model complexity while maintaining high performance
\section*{Limitations}
While our proposed approach shows promise in advancing multilingual TTS synthesis, there are several limitations that must be acknowledged. Addressing these challenges will be crucial for enhancing the robustness and applicability of our methods across a wider range of languages and use cases. The key limitations are as follows:
\begin{itemize}
    \item  The performance of hypernetworks and adapters can vary greatly depending on the hyperparameters used. Adjusting these settings for each language and task is often a complex and time-consuming process that requires significant computational resources.
    \item Languages such as Russian and Greek use scripts that differ from the Latin alphabet, like Cyrillic and Greek scripts, respectively. These scripts have unique rules for how letters and sounds are represented. The current PETL methods might not fully address these differences, resulting in lower quality speech synthesis for these languages.
    \item Symbolic languages, such as Chinese and Japanese, have unique linguistic elements like Chinese logograms and Japanese kana, as well as complex grammatical structures in languages like Russian and Greek. The proposed architecture in its current form can struggle to handle these diverse features effectively, which means modification to these adaptation techniques are needed to improve performance.
\end{itemize}

\section*{Potential Risk}
While our research aims to advance multilingual TTS technology, it is crucial to acknowledge the potential risks associated with such systems. We will discuss some associated risks as follows :
\begin{itemize}
    \item Malicious Use and Disinformation: The ability of TTS systems to generate highly realistic speech could be used to create disinformation. This could lead to the spread of false information, manipulation of opinion, and erosion of trust in digital content.

    \item Our research utilizes the publicly available CSS10 dataset, however utilization of personalized data to adapt these models can have the risk of privacy violations. Therefore it is important to follow best data management practices that do not inadvertently compromise privacy.


    \item TTS systems are vulnerable to adversarial attacks where small perturbations to the input can lead to significant changes in the output. Although the proposed framework is robust to noise, the necessary security measures and continuous testing of the system against potential attacks can enhance resilience.
\end{itemize}

\section*{Ethical Considerations}
The TTS system could be used to produce misleading or harmful content. For instance, synthesized speech could be exploited to create fake audio recordings that mimic real individuals, potentially leading to misinformation or fraud. Additionally, the accessibility of TTS technology might raise concerns about the unauthorized use of voices, infringing on personal privacy and intellectual property rights. 

\bibliography{custom}




\end{document}